\pdfoutput=1
\documentclass[11pt]{article}

\usepackage[]{EMNLP2022}

\usepackage{times}
\usepackage{latexsym}
\usepackage{algorithm}
\usepackage{algpseudocode}
\usepackage{amsmath}
\usepackage{graphics}
\usepackage{amsthm}
\usepackage{bm}
\usepackage{subfigure}
\usepackage{epsfig}
\usepackage{multirow}
\usepackage{amsmath}
\usepackage{booktabs}

\newtheorem{definition}{Definition}[section]

\usepackage[T1]{fontenc}
\usepackage[utf8]{inputenc}

\usepackage{microtype}
\DeclareMathOperator*{\argmax}{arg\,max}
\DeclareMathOperator*{\argmin}{arg\,min}

\newcommand{\adding}[1]{\textcolor{red}{}}

\title{Syntax-guided Localized Self-attention 
by Constituency Syntactic Distance}

\author{
Shengyuan Hou\textsuperscript{\rm 1}\thanks{$^{*}$ Equal contribution.}\qquad 
Jushi Kai\textsuperscript{\rm 1}$^{*}$\qquad 
Haotian Xue\textsuperscript{\rm 1}$^{*}$ \\
\textbf{Bingyu Zhu\textsuperscript{\rm 2}\qquad 
Bo Yuan\textsuperscript{\rm 2}\qquad 
Longtao Huang\textsuperscript{\rm 2}\qquad 
Xinbing Wang\textsuperscript{\rm 1}\qquad 
Zhouhan Lin\textsuperscript{\rm 1}\thanks{$\;\,$ Zhouhan Lin is the corresponding author.}} \\
\textsuperscript{\rm 1} Shanghai Jiao Tong University\qquad
\textsuperscript{\rm 2} Alibaba Group \\
\texttt{\{hsyhwjsr,json.kai,xavihart\}@sjtu.edu.cn} \\
\texttt{\{zhubingyu.zby,qiufu.yb,kaiyang.hlt\}@alibaba-inc.com} \\
\texttt{lin.zhouhan@gmail.com}
}
 
\begin{document}
\maketitle

\begin{abstract}
%The Transformer model with self-attention gives rise to revolutionary changes in a variety of seq2seq tasks such as machine translation. 
Recent works have revealed that Transformers are implicitly learning the syntactic information in its lower layers from data, albeit is highly dependent on the quality and scale of the training data. However, learning syntactic information from data is not necessary if we can leverage an external syntactic parser, which provides better parsing quality with well-defined syntactic structures.  
This could potentially improve Transformer's performance and sample efficiency.
In this work, we propose a syntax-guided localized self-attention for Transformer that allows directly incorporating grammar structures from an external constituency parser. It prohibits the attention mechanism to overweight the grammatically distant tokens over close ones. 
Experimental results show that our model could consistently improve translation performance on a variety of machine translation datasets, ranging from small to large dataset sizes, and with different source languages. \footnote{Our code is available at \url{https://github.com/LUMIA-Group/distance_transformer}} 
 %\cite{tenney2019bert}
% Comparisons with the Transformer baseline model validate the effectiveness of constituency syntactic information for learning self-attention. 
% It induces a \textit{syntactic local range} for every token, which marks the range of the grammatically closer tokens, and prohibits the attention mechanism to overweight the tokens outside of the range over the ones inside. 
% We first serialize constituency grammar through the syntactic distance mechanism, and then explicitly incorporate it by selecting several attention heads as ``grammar-aware heads'', in which the attention ranges of each token are individually modulated according to their grammatical roles. It prohibits the attention mechanism to overweight the grammatically distant tokens over close ones. 
% However, self-attention lacks inductive bias information, which hinders the learning of textual syntactic structures, especially in low-resource scenarios. Otherwise, the availability of a syntactic parser makes it easier to obtain the grammar structure of textual data. 
% into self-attention by defining an attention mask with a locality pattern generated by syntactic distance, which is a serialized representation of the constituency grammar tree.
\end{abstract}
\begin{figure*}[!t]  \label{syntactic_distance}
\begin{center}  
\subfigure[Constituency Grammar Tree]{   \includegraphics[width=0.33\linewidth]{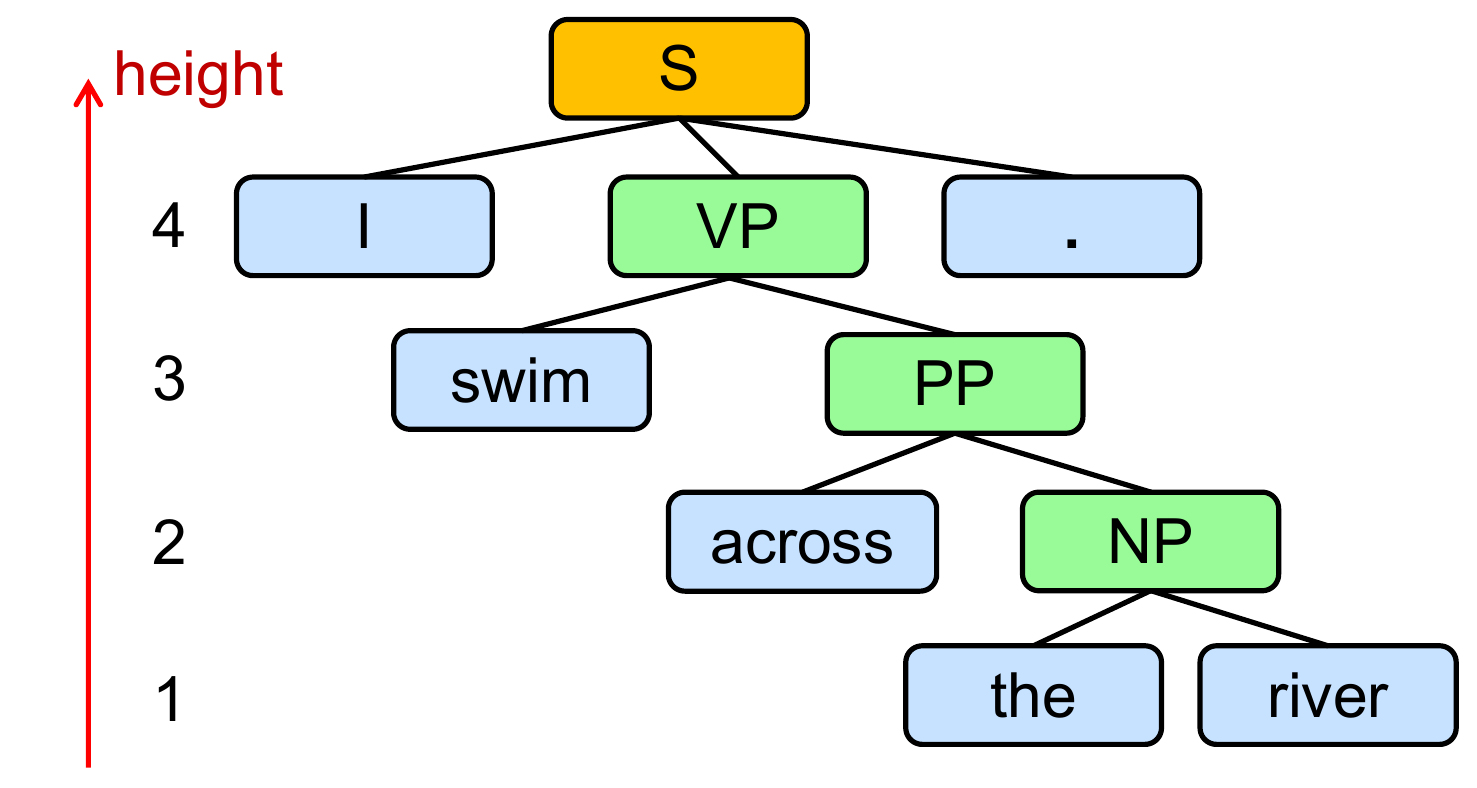}  
\label{constituency}
}  
\subfigure[Syntactic Distance]{   \includegraphics[width=0.36\linewidth]{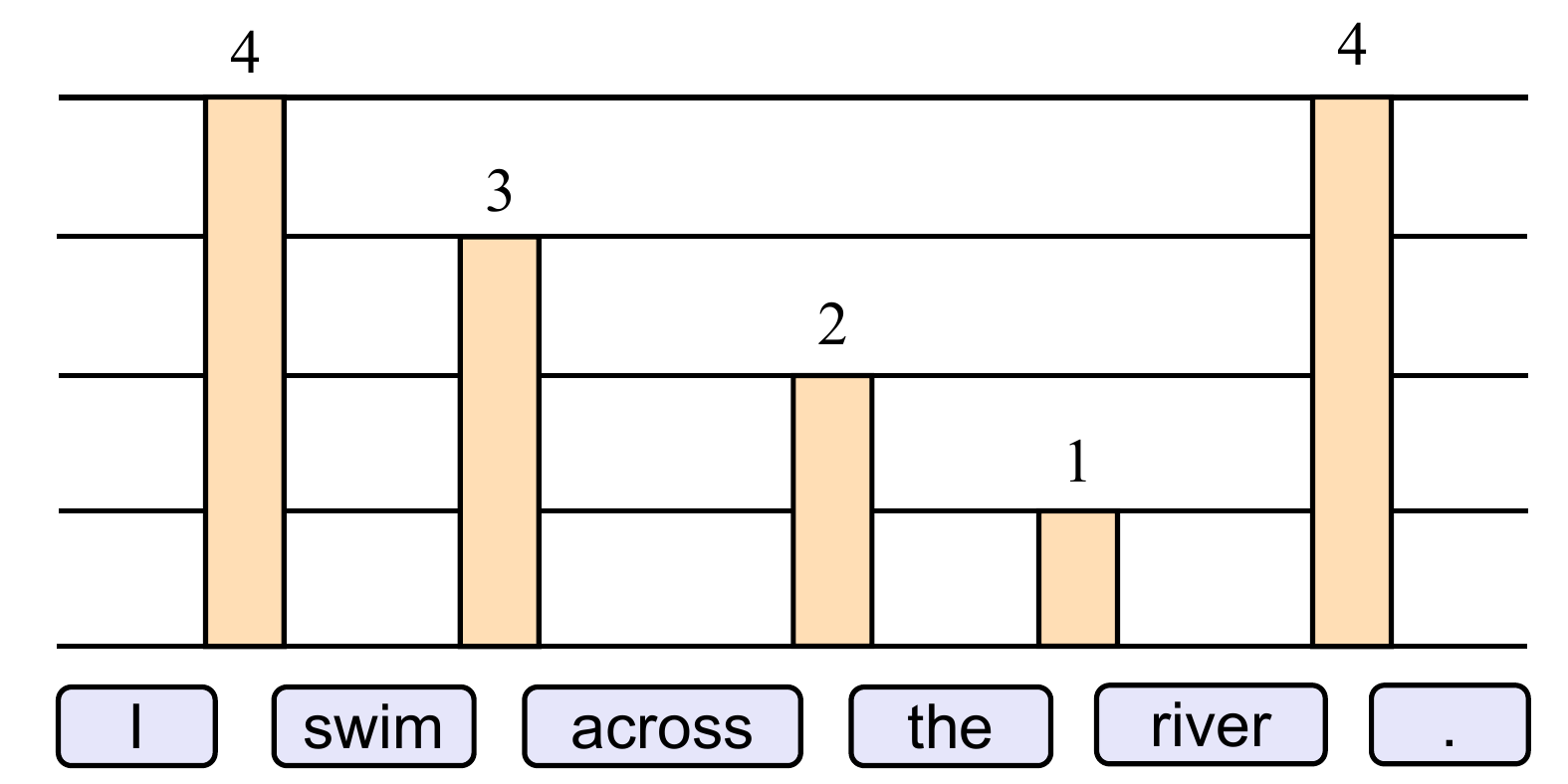}
\label{Syntactic}
}  
\subfigure[Syntactic Local Range]{   \includegraphics[width=0.26\linewidth]{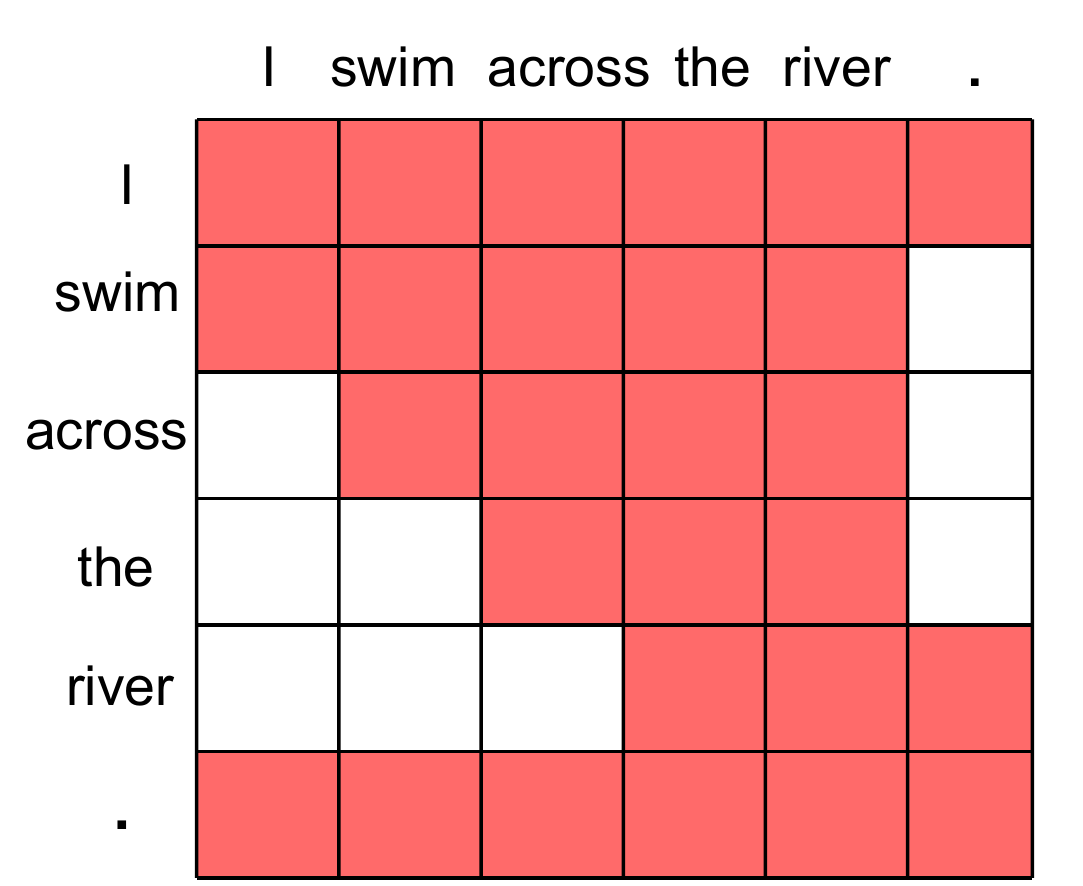}
\label{attention_mask}
}  
\caption{(a) The constituency tree for the example sentence "I swim across the river.". (b) Its syntactic distances. % The syntactic distance is a sequence of scalars which represents the merging/splitting order of tree. The highest distance values lies between \textit{I swim} and \textit{river .}, so the root node is split into "I","swim across the river" and "." at first. The distance in "the river" is the smallest so split of "the river" takes the last step. 
(c) The attention mask reflecting the syntactic local ranges of each word %defined by Definition \ref{rule1}.
For example, rather than attending to the whole sequence, "across" encourages attention towards % syntactically related tokens
\textit{swim}, \textit{the} and \textit{river} while suppresses the others.}  
%\vspace{-0.8cm}
\end{center}  
\end{figure*}
\section{Introduction}
Although Transformer doesn't have any inductive bias on syntactic structures, some studies have shown that it tends to learn syntactic information from data in its lower layers \cite{tenney-etal-2019-bert,goldberg2019assessing,jawahar-etal-2019-bert}. Given the pervasiveness of syntactic parsers that provides high quality parsing results with well-defined syntactic structures, Transformers may not need to re-invent this wheel % by learning low-level syntactic structures through data 
if grammar structures could be directly incorporated into it.

Prior to Transformer \cite{vaswani2017attention}, earlier works have demonstrated that syntactic information could be helpful for various NLP tasks. For example, \citet{DBLP:conf/acl/LevyG14} introduced dependency structure into word embeddings, and \citet{chen-etal-2017-improved} uses Tree-LSTMs to process the grammar trees for machine translation.  

More recently, dependency grammar has been successfully integrated into Transformer in various forms. \citet{strubell-etal-2018-linguistically} improves semantic role labelling by restricting tokens to only attend to its dependency parent. \citet{DBLP:conf/aaai/0001WZDZ020} modifies BERT \cite{devlin-etal-2019-bert} for named entity recognition as well as GLUE tasks \cite{wang-etal-2018-glue} by adding an additional attention layer that allows every token to only attend to its ancestral tokens in the dependency parse tree. \citet{bugliarello-okazaki-2020-enhancing} improves machine translation by constructing the attention weights from dependency tree, while \citet{li-etal-2021-improving-bert} masks out distant nodes in the dependency tree from attention. 

\textcolor{black}{While the dependency grammar demonstrates the relation between nodes, the constituency grammar focuses more on how a sentence is formed in a merging way block by block. Constituency grammar contains more information about the global structure of a sentence in a hierarchical way, which we think will greatly improve global attention mechanism like self-attention in Transformers.}
Since constituency grammar doesn't directly reflect grammatical relations between words and introduces new constituent nodes, integrating it into Transformer becomes less obvious. %, especially for the attention weights in self-attention. 
\citet{ma-etal-2019-improving} explores different ways of utilizing constituency syntax information in the Transformer model, including positional embedding, output sequence, etc. \citet{yang-etal-2020-improving-neural} uses dual encoders to encode both source text and template yielded by constituency grammar, at a cost of introducing a large amount of parameters. 

% Syntactic information is found to be helpful to enhance the performance of various NLP tasks \cite{chen2017improved, eman-dependency,li2020improving,DBLP:conf/aaai/0001WZDZ020,yang2020improving}. 

% Incorporating syntactic structures into the original self-attention mechanism can help bring linguistic inductive bias into the learning process, which will focus more on task-specific learning rather than structural bias.

% The easiest way to incorporate structural information into self-attention mechanism is to generate a syntactic-related attention mask, and then use it to affect original self-attention weight. Many methods adopt structural pattern masks from dependency-based syntax \cite{eman-dependency, wang-dependency,li2020improving}, and some work uses one specific layer of constituency tree as a linear soft template to supervise the output of the model \cite{yang2020improving}. 

% But it still remains blank about how to generate the syntactic mask directly from the integrated tree-based constituency grammar, using which we can bring explicit syntactic information into self-attention structure.

In this work, we propose a syntax-guided localized self-attention that effectively incorporates constituency grammar into Transformers, without introducing additional parameters. We first serialize constituency trees through syntactic distance \cite{shen-etal-2018-straight}, and then select several attention heads as grammar-aware heads in which the attention ranges of each token are individually modulated according to their grammatical roles. the modulated attention ranges are named as \textit{syntactic local ranges}, which prohibits the attention mechanism to overweight the grammatically distant tokens over close ones. Experimental results show that our model could consistently improve translation performance on a variety of machine translation datasets, ranging from small to large dataset sizes, and with different source languages.

\section{Preliminary: Syntactic Distance}
\subsection{Definition}
Syntactic distance \cite{shen-etal-2018-straight} is a serialized vector representation of constituency grammar tree (Fig \ref{constituency}) that is defined as:
\begin{definition} (\textbf{Syntactic Distance})
Given a length $n$ sentence $S=(t_1,...,t_n)$ and its constituency grammar tree $T$, in which the height of the lowest common ancestors of any pair of tokens $t_i, t_j$ is noted as $h_j^i$. The syntactic distance $D=(d_1,...,d_{n-1})$ of this sentence could be any vector of scalars with length $n-1$, which satisfies: $\forall i,j \in [1, n-1], $
\begin{equation} \label{eq1}
    \text{sign}(d_i-d_j) = \text{sign}(h_{i+1}^{i} - h_{j+1}^{j})
\end{equation}
  % All the scalars in $D$ are natural numbers.
\end{definition}
Intuitively, syntactic distance $D$ keeps the same ranking order as the sequence of $(h_2^{1}, h_3^2, ..., h_{n}^{n-1})$, in which $h_{i+1}^{i}$ is the height of the lowest common ancestors between pairs of consecutive words in the sentence (See Fig. \ref{Syntactic}). % The smaller $d_i$ is, the earlier two constituents in the neighborhood are merged. From another perspective, the larger $d_i$ is, the earlier two neighboring constituents are divided.
 
\subsection{Generation of Syntactic Distance}
\label{syntactic_generation}
The syntactic distance could be generated by recursively spliting the constituency tree in a top-down manner. According to the merging order of constituency syntactic tree, for any subtree T, the subtrees rooted by every child node of T must be constructed at first. Therefore, the merging of all of T's child nodes must take place  afterwards. The syntax distance in all the subtrees of T can be calculated first, and then the maximum distance value plus 1 is the current merging distance order.

During preprocessing, the syntactic distance is calculated on different datasets respectively. For each sentence, we first concatenate all BPE word segmentations, then analyze the syntax tree structure using the Stanford corenlp toolkit \cite{manning-EtAl:2014:P14-5}, and calculate the syntactic distance according to the algorithm in Algorithm \ref{distance_generation}. When filling in the syntactic distance between words in BPE word segmentation, the lowest value is 0, and finally all syntactic distances are added by 1 to become a syntactic distance vector with a minimum value of 1. In the case of multiple sentences, we generate the syntactic distance of each sentence respectively, and then fill in the maximum 999 between different sentences, indicating that all sentences are merged at last.

\begin{algorithm}
\caption{Syntactic Distance Generation}
\label{distance_generation}
\begin{algorithmic}[1]
\Require Constituency Grammar Tree $T$
\Ensure Syntactic Distance $D=(d_1,...,d_n)$
\Function{distance}{root}
\State $\bm{d}\leftarrow [\ ]$
\If{root is not leaf} //Node can be split
    \State $d\_set\leftarrow \{\ \}$
    \State $maxd\leftarrow 0$
    \For{c in root's child} 
    \State $\bm{c_d}\leftarrow Distance(c)$
    \State $maxc\leftarrow \max(c_{d,1},...,c_{d,|\bm{c_d}|})$
    \State $d\_set\leftarrow d\_set \cup \{\bm{c_d}\}$
    \State $maxd\leftarrow \max (maxd, maxc)$
    \EndFor
    \State $nd\leftarrow maxd+1$
    \State $\bm{d}\leftarrow [\bm{c_{d1}};nd;\bm{c_{d2}};...;nd;\bm{c_{d|d\_set|}}]$ 
\EndIf
\State \Return $\bm{d}$;
\EndFunction
\end{algorithmic}
\end{algorithm}

% For some root node, since its children are split later than itself, the distance between its subnodes should be larger than distance entailed by its subtrees. 

% \begin{algorithm}[h]    
% \caption{Syntactic Distance Calculation}    
% \label{TreetoDistance}    
% \KwData{Constituency Syntax Tree $T$}      

% \KwResult{Syntactic Distance $D=(d_1,d_2,...,d_n)$}   
% % \begin{algorithmic}[1]
% \State Generate_Distance(root);
% \State Mer;      
% \Repeat        
% \State compute gradient directions $g_k=\bigtriangledown f(x_k)$;        
% \State compute Polak-Ribiere parameter $\beta_k=\frac{g_k^{T}(g_k-g_{k-1})}{\parallel g_{k-1} \parallel^{2}}$;        
% \State compute the conjugate directions $d_k=-g_k+\beta_k d_{k-1}$;        
% \State compute the step size $\alpha_k=s/\parallel d_k \parallel_{2}$;      
% \Until{($f(x_k)>f(x_{k-1})$)}   
% % \end{algorithmic}  
% \end{algorithm}

\section{Method}\label{sec:method}
%For any token in a sentence, it have its syntactic local pattern related to its position in the constituency tree. We regard it as the perception range. 
We are going to present a form of local self-attention that dynamically controls each word's attention range according to its syntactic role in the sentence (See Fig. \ref{attention_mask}). 
Attention heads that incorporate this localized self-attention could significantly outweight the grammatically close tokens over distant ones, thus incorporate the syntax information as prior knowledge.

% In this section, we will firstly define the syntactic local range, and then show how to induce them from the syntactic distances, as well as how it is incorporated into vanilla self-attention.

\subsection{Syntactic Local Range}
It is widely believed that in a grammar tree, sibling words within a certain constituent is more correlated than words that are more distant \cite{klein-manning-2003-accurate}. For example, in Fig. \ref{constituency}, for the word \textit{the}, the word \textit{river} locates to its right is more related than the word \textit{I}. More generally, we define a \textit{syntactic local range} for each of the words to mark a set of adjacent words as more syntactically correlated to it. 
For a given token $t$, we consider the words that are 1) direct siblings of $t$; or 2) left/right siblings of the constituent that has $t$ on its left/right boundary respectively, as syntactically more correlated to $t$. 
Note that, depending on the structure of the grammar tree, the syntactic local range to the left of the word could be different with that to the right. 
Formally, we have the following definitions. 

% Considering the formation of a constituency tree as a bottom-up merging of tokens: atomic tokens are gradually merged into constituents with more broad and abstract meaning. Intuitively, tokens can attend more to its syntactic related tokens. On the basis of this, in the tree-level, 
% we define the \textbf{Syntactic Local Range} as follows.
\begin{definition} (\textbf{Syntactic Local Range}) For a given token $t$, let $f$ be its parent node in the constituency tree. Syntactic local range is defined as the concatenation of two ranges residing on $t$'s left and right sides, corresponding to its pre-text and post-text directions, respectively. 

For the pre-text direction,  
\begin{itemize}
    \item If $t$ is not the leftmost child of $f$, then $t$'s syntactic local range on its left side starts at $f$'s leftmost child, and stops at $t$.

    \item If $t$ is the leftmost child of $f$, back-trace its ancestors to find a nearest constituent $v$ where $v$ is not the leftmost child of its parent $w$. Then $t$'s syntactic local range on its left side starts at $w$'s leftmost child and stops at $t$.
\end{itemize}

Similar rules apply for the pos-text direction except that the left/right directions are altered accordingly.
\label{rule1}
\end{definition}

% Now that we have clearly defined the range within which we want to put more attention weights on, next we will show how to compute the ranges for every tokens from syntactic distances.

% It's intuitive that when some token lies in the interior of its parent's subtree, the range it can attend to is only all of its siblings in locality. When it's on the edge, it could attend to farther neighborhood tokens representing its parent's subtree(Fig \ref{syntactic_distance}).   

\subsection{Inducing Syntactic Local Range from Syntactic distance}
% The syntactic local ranges for all the tokens in the sentence could be induced by syntactic distances. 
Intuitively, for a given token $t$, its syntactic local range keeps stretching on both pre-text and post-text directions until it hits a distance that is larger than $t$'s on the corresponding side. We can represent all the syntactic local ranges for every words in a sentence as a masking matrix consisting of 0s and 1s (c.f. Fig. \ref{attention_mask}), and compute it through Algorithm \ref{pattern_generation}. It can be proved that the ranges generated by Algorithm \ref{pattern_generation} is identical to the syntactic local ranges defined by Definition \ref{rule1}:

% On pre/pos-text direction, token $t_i$ is attended to the first left/right token with a bigger neighboring syntactic distance. We formulate this in Algorithm \ref{pattern_generation}.

% \begin{definition} Given a sentence $S$ and its tokenization $(t_1,...,t_n)$. Assume that the calculated syntactic distance is $D=(d_1,...,d_{n-1})$. 
% \quad  
% \begin{itemize} 
%     \item For a token $t_i$, we begin from its left syntactic distance $d_{i-1}$, the leftmost token id $b_l$ which $t_i$ can attend to is :
%     $b_l = \argmax\limits_{j < i-1,d_j > d_{i-1}}j+1$
%     \item For a token $t_i$, we begin from its right syntactic distance $d_{i}$, the rightmost token id $b_r$ which $t_i$ can attend to is :
%     $b_r = \argmin\limits_{j > i,d_j > d_{i}} j$
% \end{itemize}
% \label{rule2}
% \end{definition}

\begin{algorithm}[t]
\caption{Inducing Syntactic Local Range}\label{pattern_generation}
\begin{algorithmic}[1]
\Require Syntactic Distance $D=d_i(i \in [1, n-1])$
\Ensure Syntactic Local Range $G$

\State $G = \textbf{0} \in \{0, 1\}^{n\times n}$

\For{i=$1$ to $n$}
\State    $b_l = \argmax\limits_{j < i-1,d_j > d_{i-1}} \{ j+1 \}$ $//$ left boundary
\State    $b_r = \argmin\limits_{j > i,d_j > d_{i}} \{ j \}$ $//$  right boundary
\State    $G[i, b_l : b_r] = 1$
\EndFor

\State \Return $G$

% \If{root is not leaf} \Comment{Node can be split}
%     \State $d\_set\leftarrow \{\ \}$
%     \State $maxd\leftarrow 0$
% \EndIf
% \State \Return $\bm{d}$;
% \EndFunction
\end{algorithmic}
\end{algorithm}

% \begin{algorithm}[h]    
% \caption{Syntactic Distance Calculation}    
% \label{TreetoDistance}    
% \KwData{Constituency Syntax Tree $T$}      

% \KwResult{Syntactic Distance $D=(d_1,d_2,...,d_n)$}   
% % \begin{algorithmic}[1]
% \State Generate_Distance(root);
% \State Mer;      
% \Repeat        
% \State compute gradient directions $g_k=\bigtriangledown f(x_k)$;        
% \State compute Polak-Ribiere parameter $\beta_k=\frac{g_k^{T}(g_k-g_{k-1})}{\parallel g_{k-1} \parallel^{2}}$;        
% \State compute the conjugate directions $d_k=-g_k+\beta_k d_{k-1}$;        
% \State compute the step size $\alpha_k=s/\parallel d_k \parallel_{2}$;      
% \Until{($f(x_k)>f(x_{k-1})$)}   
% % \end{algorithmic}  
% \end{algorithm}

\begin{proof}
Consider the token $t_i$ in the sequence $(t_1,t_2,...,t_n)$ with the syntactic distances $(d_1,d_2,...,d_{n-1})$. 

For the pre-text direction, let $t_i$'s syntactic local range starts at $t_l$. 
According to Definition \ref{rule1}, $t_l$ and $t_i$ share a lowest common ancestor $w$.  Let the height of $w$ be $h_w$.  
\begin{itemize}
    \item[1.] Since $t_l$ is the leftmost child of $w$, we have $h_l^{l-1} >= h_w$.
    \item[2.] Since $t_i$ is either the immediate child of $w$, or the leftmost child of a constituent $v$ who is the immediate child of $w$, we have $h_i^{i-1} = h_w - 1$.
    \item[3.] For all tokens between $t_l$ and $t_i$ inclusively, $w$ should also be their common ancestor. Thus we have $\forall j \in [l+1, i], h_j^{j-1} <= h_w -1$.  % $t_{l}, t_{l+1}, ..., t_{i-1}$, 
\end{itemize}
Comprehensively, we have 
\begin{equation}
    \forall j \in [l+1, i-1], h_l^{l-1} > h_i^{i-1} >= h_j^{j-1}
\end{equation}
According to Eq. \ref{eq1}, we could derive that 
\begin{equation}
    \forall j \in [l, i-2], d_{l-1} > d_{i-1} >= d_{j}
\end{equation}
Thus validates line 3 in Algorithm \ref{pattern_generation}.

Symmetrically, the proof for the post-text direction is obvious. % Therefore, attention range defined by Definition \ref{rule1} is identical to that calculated by Algorithm \ref{pattern_generation}.
\end{proof}

Specifically, we first define the syntactic local pattern in a constituency tree in Definition \ref{rule1}, which is explicit and easy to understand. However, using syntactic distance can facilitate the computing of syntactic local pattern since it can be well paralleled. Therefore we turn to an alternative expression of Definition \ref{rule1} using syntactic distance, and propose Algorithm \ref{pattern_generation} to generate mask from syntactic distance, which is also proved to be identical to the definition of syntactic local pattern.

% Define the parent of node $t_i$ as $p_i$, rooting a sub-tree $T_i$. Syntactic distance on the left of $t_i$ is $d_{i-1}$.
% \begin{itemize}
%     \item[1.] If $t_i$ is not $T_i$'s leftmost child, we note the leftmost one as $l_{T_i}$. Assume that the leftmost leaf node in subtree rooted by $l_{T_i}$ is $\text{leaf}_{p_i}$. Since $t_i$ and its siblings are merged in the same turn and the sub-trees rooted by $t_i$'s siblings are constructed earlier, it can be concluded from Algorithm \ref{distance_generation} that $d_{i-1} \geq d_j(\text{leaf}_{p_i}\leq j <i-1)$. Since subtree rooted by $p_i$ is merged later than nodes within this subtree, we have $d_{\text{leaf}_{p_i}-1} > d_{i-1}$. According to Algorithm \ref{pattern_generation}, $\text{leaf}_{p_i}$ is the leftmost token $t_i$ can attend to.
%     \item[2.] If $t_i$ is $T_i$'s leftmost child, then we move upward to find the first node $A_i$ which is not the leftmost child of its parent. Note the leftmost sibling of $A_i$ as $l_{A_i}$, and the leftmost leaf node of parental subtree $p_{A_i}$ as $\text{leaf}_{p_{A_i}}$. Since the construction of all $A_i$'s left siblings is earlier than their merge, we have $d_{i-1} \geq d_j (\text{leaf}_{p_{A_i}} \leq j < i - 1)$ . Since subtree rooted by $p_{A_i}$ is merged later than nodes within this subtree, we have $d_{i-1} < d_{\text{leaf}_{p_{A_i}}-1}$. According to Algorithm \ref{pattern_generation}, the leftmost token $t_i$ can attend to is $\text{leaf}_{p_{A_i}}$.

\subsection{Syntax-guided Self-Attention}
Given the masking matrix $G$, we can incorporate it into self-attention through masked softmax to get a syntax-guided self-attention:
\begin{equation}
    \text{Attn}(Q, K, V, G) = \text{S}_{m} \left ( G, \frac{QK^{T}}{\sqrt{d_k}} \right ) V
\end{equation}
\begin{equation}
   \text{S}_{m} \left( M,X \right) = {\left [ \text{S}\left (\frac{m_{ji} e^{x_{ji}}}{\sum_{k=1}^{n} m_{jk} e^{x_{jk}}} \right ) \right ] }_{j=[1,...,n]}
\end{equation}
where $\text{S}_{m}(\cdot)$ and $\text{S}(\cdot)$ are masked softmax and softmax, respectively. And $m_{ji}$ and $x_{ji}$ are the element in the $j$-th row and $i$-th column of the $M$ and $X$ matrices. $\left [ \cdot \right ]_{j=[1,...,n]}$ corresponds to vertically stack the elements yielded by the function inside it, whose input $j$ being $1,...,n$ respectively. 

% \subsection{Syntactic Mask}
Moreover, since we want to encourage the attention within the syntactic local range and suppress those outside of it, rather than zeroing out the attention weights outside of the syntactic local range. To this end, we would still like the model to attend on tokens outside of the range. Thus, rather than using a crispy boundary, i.e., elements in $G$ are either $0$ or $1$, we smooth it through the following steps. 

% We implement Algorithm \ref{pattern_generation} by parallel matrix operation. 

First we define the \textit{soft comparison} between $d_i$ and $d_j$ as 
\begin{equation}\label{smooth-alpha}
    \alpha_{i}^{j}=\frac{\text{tanh}((d_i- d_j) / \tau) + 1}{2}
\end{equation}
which yields $0$ when $d_i \ll d_j$ and gradually increases to $1$ as $d_i$ increases, with $\tau$ controlling the softness.

% First we transform the magnitude relationship of syntactic distances into a continuous value $\alpha_{i}^{j} \in (0, 1)$. $\alpha_{i}^{j}$ is close to 1 when $d_i>d_j$.

% According to Algorithm \ref{pattern_generation}, when attending to left/right, the token needs to find the first left/right larger value of syntactic distance to define the range. 
We can simulate the attention range found by Algorithm \ref{pattern_generation} by multiplying $\alpha$ from the central token to both sides. i.e., entries in $G$ can be approximated by % Then we can get the syntactic mask value $G_{ij}$ which is correspondent to the attention weight from $t_i$ to $t_j$:
\begin{equation}
    G_{ij}=\prod_{j\leq t \leq i-1}\alpha_{t}^{i-1} , j < i-1
    % \vspace{-0.25cm}
\end{equation}
To initiate the continued product, mask values on the two secondary diagonals in $G$ are set to 1. 

We then use this syntactic-guided attention mask to augment some of the attention heads in the vanilla Transformer model, thus incorporates constituency grammar information into Transformer.

% in order to prevent from destroying Transformer's ability of learning long-range dependency.  

\section{Experiments}
\subsection{Datasets and Experiment Settings}
We implement our model based on Fairseq \cite{ott-etal-2019-fairseq} toolkit %\footnote{https://github.com/pytorch/fairseq}. 
The Stanford CoreNLP parser %\footnote{https://stanfordnlp.github.io/CoreNLP/}
is utilized to generate constituency grammar tree, which is further used to calculate syntactic distance. All distance values are natural numbers with a minimum value of 1 and distances between sentences are set to 999 by default.  

We evaluate our model on machine translation tasks of different languages. We use IWSLT-14 German-English on both directions (De2En and En2De), NC11 German-English on both directions, ASPEC from Chinese to Japanese (Ch2Ja), and WMT14 from English to German. Unless otherwise noted, we only modify the first encoder layer, with 3 heads for IWSLT14-En2De and NC11-En2De, 4 heads for NC11-De2En and IWSLT14-De2En, 2 heads for ASPEC-Ch2Ja and  WMT14-En2De. % Other attention heads are taken as original self-attention.
Detailed settings and available in Appendix \ref{training}.

\subsection{Experimental Results}

\begin{table}[t]
	\centering  % 显示位置为中间

	\begin{tabular}{ccc} 
		\toprule
		datasets & Transformer & Ours \\
		\midrule
		IWSLT14-De2En & 34.56 & \textbf{35.74} \\
		IWSLT14-En2De & 28.17 & \textbf{29.28} \\
		NC11-De2En & 26.83 & \textbf{27.67} \\
		NC11-En2De & 25.26 & \textbf{26.19} \\
		ASPEC-Ch2Ja & 47.77 & \textbf{48.34} \\
		WMT14-En2De & 27.3 & \textbf{28.48} \\
		
		\bottomrule
	\end{tabular}
	\caption{Test BLEU score on five datasets.}
	\label{Total-0}  
% \vspace{-1.0em}
\end{table}

We compare our syntactic-guided model with Transformer \cite{vaswani2017attention} measuring by BLEU score\footnote{https://pypi.org/project/sacrebleu/}. All of the results are averaged over 5 independent runs on different seeds and the improvements have statistical significance (p<0.01).   %checkpoints selected by validation loss on five different random seed initializations (1$\sim$5) and have statistical significance (p<0.01). 
For the large WMT dataset, we average the test results over the last 5 checkpoints in a single run due to computation limits. Our model outperforms Transformer baseline model on all six machine translation datasets. Our model can achieve up to 1.18 improvements on these test beds. Detailed results are shown in table \ref{Total-0}.

\subsection{Ablation Study}

\begin{figure}[t]
    \includegraphics[width=0.9\linewidth]{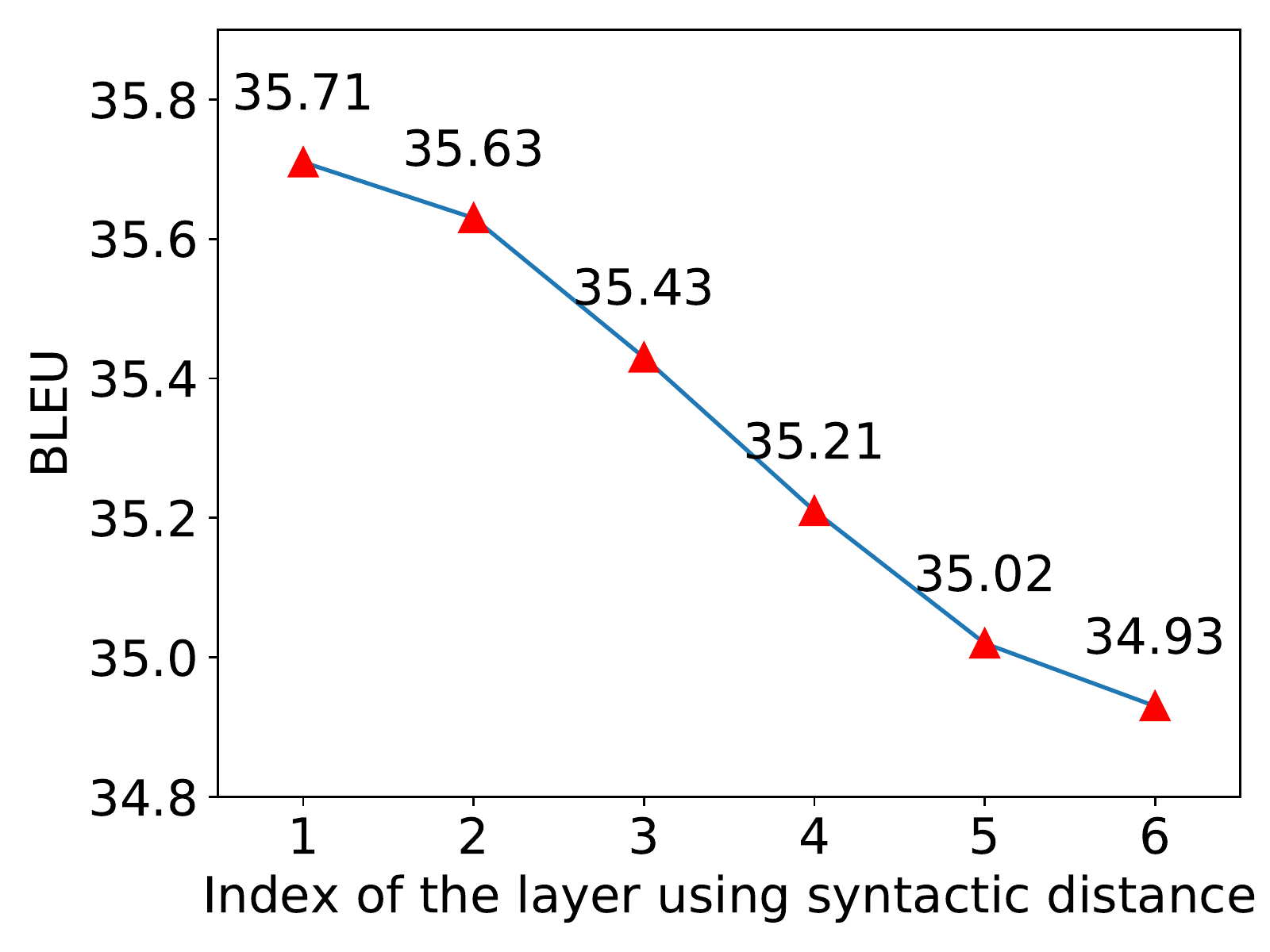}
    \caption{Layer-wise for IWSLT14-De2En}
    \label{layer-wise}
\end{figure}

\begin{table}[t]
	\centering  % 显示位置为中间

	\begin{tabular}{ccccc} 
		\toprule
		\multirow{2}{*}{models} &
		\multicolumn{2}{c}{IWSLT14} &
		\multicolumn{2}{c}{NC11} \\
		\cline{2-5} & De2En & En2De & De2En & En2De \\
		\midrule
		baseline & 34.56 & 28.17 & 26.83 & 25.26 \\
		1-head & 35.65 & 29.20 & 27.58 & 25.87 \\
		2-head & 35.68 & 29.21 & \textbf{27.67} & 25.99 \\
		3-head & 35.71 & \textbf{29.28} & 27.56 & \textbf{26.19} \\
		4-head & \textbf{35.74} & 29.23 & \textbf{27.67} & 26.08 \\
		\bottomrule
	\end{tabular}
	\caption{Head-wise for IWSLT14 and NC11.}
	\label{head-wise}  
\end{table}

We apply syntactic distance in different encoder layers of Transformer to get best utilization of syntactic distance. The results are shown in Fig. \ref{layer-wise}, where "x-layer" means that only 3 attention heads of x-th encoder layer use syntactic information. The model performs best with syntactic distance used in the first layer. Moreover, the deeper layer we add syntactic distance, the worse performance the model has. It can be concluded that top layers focus more on long range dependency of data.

% \begin{table}[htbp]
% 	%\setlength{\abovecaptionskip}{0.cm}
% 	%\setlength{\belowcaptionskip}{-0.cm}
% 	\centering  % 显示位置为中间

% 	\begin{tabular}{cc} 
% 		%\setlength{\tabcolsep}{20mm}
% 		\toprule
% 		models & IWSLT14-De2En \\
% 		\midrule
% 		1-layer & \textbf{35.71}  \\
% 		2-layer & 35.63  \\
% 		3-layer & 35.43  \\
% 		4-layer & 35.21  \\
% 		5-layer & 35.02  \\
% 		6-layer & 34.93  \\
% 		\bottomrule
% 	\end{tabular}
% 	\caption{Layer-wise for IWSLT14-De2En.}
% 	\label{layer-wise}  
% % \vspace{-1.0em}
% \end{table}

We also studied effects of different heads in using syntactic distance, as shown in table \ref{head-wise}. "x-head" means we only use syntactic distance in x heads of the first encoder layer. We found that the utilization of different extent does not lead to great difference in performance.

\section{Conclusion}
\label{sec:bibtex}
In this work, we introduce constituency grammar into Transformer by using syntactic distance to guide the self-attention mechanism. We implement the complete pipeline of fusion by converting constituency tree to syntactic distance and finally generating syntactic local pattern. Experiments show that our method could consistently improve the performance of Transformer on a variety of machine translation tasks. We also excavate that the syntactic-guided attention achieve the best effectiveness in the bottom layer. Future work could expand to other NLU tasks and larger-scale pretrained models.

\section*{Limitations}
Although the constituency-based syntactic local patterns implemented by syntactic distance appear to be advantageous for self-attention learning in seq2seq architecture, there are still some directions for further improvement.

Firstly, the constituency grammar information is generated explicitly using external parser, whereas most low-resource languages like Southeast Asian family lack sophisticated parsing toolkit. Specially consituency syntax tree bank is far rarer compared with dependency tree bank. Due to this, our experiments do not include low resource languages. To solve this problem, incorporating syntactic information implicitly without any annotation is a heated topic with great prospect. We will further explore this direction.

Also, most parsers are based on statistical methods so that generated grammar tree often entails noise and uncertainty. These hinder the wide applications of our method. A more modern parser like the Berkeley Neural parser with Bert or Roberta etc. embeddings would do better probably. It would be interesting to see whether the effect on the BLUE scores of using a parser has a higher accuracy that could presumably then provide the basis for an improved incorporation of syntactic representation.

Secondly, the utilization position of syntactic mask is a tough problem when it is manually tuned, especially on large model configurations and large datasets. One alternative is to learn the usage of syntactic attention mask adaptively and softly, by which every attention head in every layer is a combination of original global attention and syntactic masked attention. The combination weight of every head is parameterized and learned throughout the training process. 

Thirdly, the pretraining and finetuning architecture has been around in recent years for its generality and excellent performance compared with the task-specific model training from scratch. It's intuitive to explore whether our method could still achieve good performance on pretraining architecture such as BERT, even though some studies find that trained on a large corpus the pretraining model has inherently entailed syntactic structural information.   

Fourthly, many studies focus on the usage of dependency syntax information to enhance the self-attention learning. The constituency grammar structure is naturally suitable for languages with strong local structure, while dependency grammar is suitable for languages with high flexibility of permutation. The difference between two kinds of syntax structure on different source languages are left for further study.  

\section*{Acknowledgements}
This work is sponsored by the National Natural Science Foundation of China (NSFC) grant (No. 62106143), and Shanghai Pujiang Program (No. 21PJ1405700). We would also like to thank the anonymous reviewers for their constructive comments and suggestions.

% Entries for the entire Anthology, followed by custom entries
\bibliography{anthology,custom}
\bibliographystyle{acl_natbib}

\appendix

\label{sec:appendix}

 \section{Training Details}
 \label{training}
We perform experiments on six machine translation datasets and use the BLEU score to evaluate our model's performance. ISWLT14 EN/DE dataset is stemmed from TED talks. NC11 EN/DE is stemmed from news commentary. ASPEC CH/JA is stemmed from scientific paper excerpt corpus. WMT14 EN/DE is stemmed from European news. Sizes of different datasets are listed in Table\ref{Dataset}.

\begin{table}[htbp]
	\centering  % 显示位置为中间
	\begin{tabular}{cccc} 
		\toprule
		datasets & Train & Valid & Test \\
		\midrule
		IWSLT14-De/En & 160239 & 7283 & 6750\\
		NC11-En/De & 238843 & 2169 & 2999\\
		ASPEC-Ch/Ja & 672315 & 2090 & 2107\\
		WMT14-En/De & 4528223 & 3000 & 3003 \\
		\bottomrule
	\end{tabular}
	\caption{Sizes of Experimental Datasets}
	\label{Dataset}  
% \vspace{-1.0em}
\end{table}

For Iwslt14 dataset preprocessing, we use Moses toolkit\footnote{https://github.com/moses-smt/mosesdecoder} for uncased word tokenization. Then we clean up the dataset, and only keep the pairs whose source-target length ratio is within 1.5 and the total length does not exceed 175. Next we use the subword NMT toolkit\footnote{https://github.com/rsennrich/subword-nmt} for BPE subword tokenization with a shared dictionary of size 10k. Finally we randomly select 5\% of training as the validation set, and merge dev2010, dev2012, tst2010, tst2011 and tst2012 into the test set. The size of the training, validation and test set are 160k/7.3k/6.7k respectively.

For NC11 dataset preprocessing, we still use Moses toolkit for uncased word tokenization. Then we clean up the dataset, and only keep the pairs whose source-target length ratio is within 1 and the total length does not exceed 80. Next we use the subword NMT toolkit for BPE subword tokenization with a shared dictionary of size 16k. Otherwise, we dropout the pairs which do not have a correct language. Validation and test set are newstest2015 and newstest2016 respectively. The size of the training, validation and test set are 239k/2.2k/3k respectively.

For ASPEC dataset preprocessing, we use Moses to extract statements. Then we use StanfordNLP-2014-01-01-segmenter\footnote{https://nlp.stanford.edu/software/segmenter.shtml} for Chinese and juman-7.0\footnote{https://nlp.ist.i.kyoto-u.ac.jp} for Japanese in word tokenization. Next we use the subword NMT toolkit for BPE subword tokenization with a Chinese dictionary of size 61k and a Japanese dictionary of size 46k. The size of the training, validation and test set are 672.3k/2k/2.1k respectively.

For WMT dataset preprocessing, subword-NMT and Moses are still ustilized for tokenization and dictionary built-up. The size of BPE shared dictionary is 40000. Non-printing characters are removed and all punctuations are normalized. The source-target length ratio and the total length are limited within 1 and 250 respectively. We sample 1\% of training data as validation set and test sets are directly downloaded.The size of training, validation and test set are 4528k/3k/3k respectively. 

Manual Tuning is performed for hyper-parameters tuning on the basis of test BLEU score for the best checkpoint. We use Adam optimizer with $\beta_1 = 0.9$, $\beta_2 = 0.98$, and $\epsilon = 10^{-8}$. We use label smoothing of value $\epsilon_{ls}=0.1$ and weight decay of 0.0001. By default, the attention-dropout is set to 0.2, dropout is set to 0.3, except that we set the attention-dropout to 0.1 in ASPEC and set the attention-dropout and dropout both to 0.1 in WMT. We take the inverse square root learning rate scheduler and set the peak learning rate as 1e-3 with 4000 steps linear warm-up and 1e-7 initial learning rate. The scaling parameter $\tau$ is set to 10. The model has a hidden dimension of 512, 6 encoder layers and 6 decoder layers with 4 attention heads per layer, except for WMT dataset with 8 attention heads per layer. Hidden layer size for FFN is set to 2048 in ASPEC and 1024 in other datasets. For IWSLT14 and ASPEC, each update has up to $8192 \times 2 \times 1$ tokens(max\_tokens$\times$gpu\_num$\times$num\_updates ). For NC11, each update has up to $8192 \times 2 \times 2$ tokens. We use 8 80G-A100 GPUs from distributed clusters to train our model on WMT14 dataset with CUDA version 11.1 and 2 40G-A100 GPUs for other datasets with CUDA version 11.2.

\begin{table}[htbp]
	\centering  % 显示位置为中间
	\begin{tabular}{ccc} 
		\toprule
		datasets & Ours & Transformer \\
		\midrule
		IWSLT14-De2En & 80min & 70min \\
		IWSLT14-En2De & 80min & 80min \\
		NC11-En2De & 80min & 80min \\
		NC11-De2En & 80min & 80min \\
		ASPEC-Ch2Ja & 12h & 12h \\
		WMT14-En2De & 9h & 8.5h \\
		\bottomrule
	\end{tabular}
	\caption{Training time of Different Datasets}
	\label{Time}  
% \vspace{-1.0em}
\end{table}

Since our incorporation of syntactic information is based on explicit calculation of syntactic distance and local attention pattern mask by external syntactic parser, we do not introduce extra parameters compared with original Transformer model. The number of parameters is 50M for configuration of WMT dataset, 99M for ASPEC dataset and 30M for the other datasets.

% As for the training time, the Transformer baseline on all of small datasets including IWSLT14 and NC11 costs about 80 minutes. Our distance enhanced model costs still 80 minutes which is identical to baseline. 

The training time of Transformer and our distance enhanced model are listed as in table \ref{Time}. Basically, our model costs about the same time for training as Transformer.

\end{document}